\documentclass[sigconf]{acmart}

\usepackage{dsfont}
\usepackage{bbm}
\usepackage{mathbbol}
\usepackage{xcolor}
\usepackage{amsmath,amsfonts}
\usepackage[capitalize]{cleveref}
\usepackage{url}
\usepackage{color,soul}
\usepackage[font=small]{caption}
\usepackage{subcaption}
\usepackage[noend]{algpseudocode}
\usepackage{afterpage}
\usepackage{tabularx}
\usepackage{booktabs}
\usepackage{arydshln}
\usepackage{url}
\usepackage[shortlabels]{enumitem}
\usepackage{bigstrut}
\usepackage{array}
\usepackage{pifont}
\usepackage{relsize}
\usepackage{cleveref}
\usepackage{xcolor}
\usepackage{fancyhdr}
\usepackage{array,multirow,graphicx}
\definecolor{darkgreen}{rgb}{0,0.5,0}

\newtheorem*{remarkCMD}{Definition CMD}
\newtheorem*{remarkTransformer}{Definition Transformer}

\AtBeginDocument{%
  \providecommand\BibTeX{{%
    \normalfont B\kern-0.5em{\scshape i\kern-0.25em b}\kern-0.8em\TeX}}}

\copyrightyear{2020} 
\acmYear{2020} 
\setcopyright{rightsretained} 

\acmConference[MM '20]{Proceedings of the 28th ACM International Conference on Multimedia}{October 12--16, 2020}{Seattle, WA, USA}
\acmBooktitle{Proceedings of the 28th ACM International Conference on Multimedia (MM '20), October 12--16, 2020, Seattle, WA, USA}
\acmPrice{} 
\acmDOI{10.1145/3394171.3413678}
\acmISBN{978-1-4503-7988-5/20/10}

\acmSubmissionID{mmfp1541}

\begin{document}

\title{MISA: Modality-Invariant and -Specific Representations for Multimodal Sentiment Analysis}

\author{Devamanyu Hazarika}
\affiliation{%
  \institution{School of Computing \\ National University of Singapore}
}
\email{hazarika@comp.nus.edu.sg}

\author{Roger Zimmermann}
\affiliation{%
  \institution{School of Computing \\ National University of Singapore}
}
\email{rogerz@comp.nus.edu.sg}

\author{Soujanya Poria}
\affiliation{%
 \institution{ISTD, Singapore University of Technology and Design}
}
\email{sporia@sutd.edu.sg}

\renewcommand{\shortauthors}{Hazarika, et al.}

\begin{abstract}
Multimodal Sentiment Analysis is an active area of research that leverages multimodal signals for affective understanding of user-generated videos. The predominant approach, addressing this task, has been to develop sophisticated fusion techniques. However, the heterogeneous nature of the signals creates distributional modality gaps that pose significant challenges.  In this paper, we aim to learn effective modality representations to aid the process of fusion. We propose a novel framework, MISA, which projects each modality to two distinct subspaces. The first subspace is modality-invariant, where the representations across modalities learn their commonalities and reduce the modality gap. The second subspace is modality-specific, which is private to each modality and captures their characteristic features. These representations provide a holistic view of the multimodal data, which is used for fusion that leads to task predictions. Our experiments on popular sentiment analysis benchmarks, MOSI and MOSEI, demonstrate significant gains over state-of-the-art models. We also consider the task of Multimodal Humor Detection and experiment on the recently proposed UR$\_$FUNNY dataset. Here too, our model fares better than strong baselines, establishing MISA as a useful multimodal framework.
\end{abstract}



\maketitle

\pagestyle{fancy}
\fancyhead{}
\fancyfoot{}

\section{Introduction}

With the abundance of user-generated online content, such as videos, \textit{Multimodal Sentiment Analysis} (MSA) of human spoken language has become an important area of research~\cite{DBLP:conf/wassa/Mihalcea12,poria2020beneath}. Unlike traditional affect learning tasks performed on isolated modalities (such as text, speech), multimodal learning leverages multiple sources of information comprising language (text/transcripts/ASR), audio/acoustic, and visual modalities. Most of the approaches in MSA are centered around developing sophisticated fusion mechanisms, which span from attention-based models to tensor-based fusion~\cite{DBLP:journals/inffus/PoriaCBH17}. 
Despite the advances, these fusion techniques are often challenged by the modality gaps persisting between heterogeneous modalities. 
Additionally, we want to fuse complementary information to minimize redundancy and incorporate a diverse set of information. One way to aid multimodal fusion is to first learn latent modality representations that capture these desirable properties. To this end, we propose MISA, a novel multimodal framework that learns factorized subspaces for each modality and provides better representations as input to fusion.

\begin{figure}[t]
    \centering
    \includegraphics[width=\linewidth]{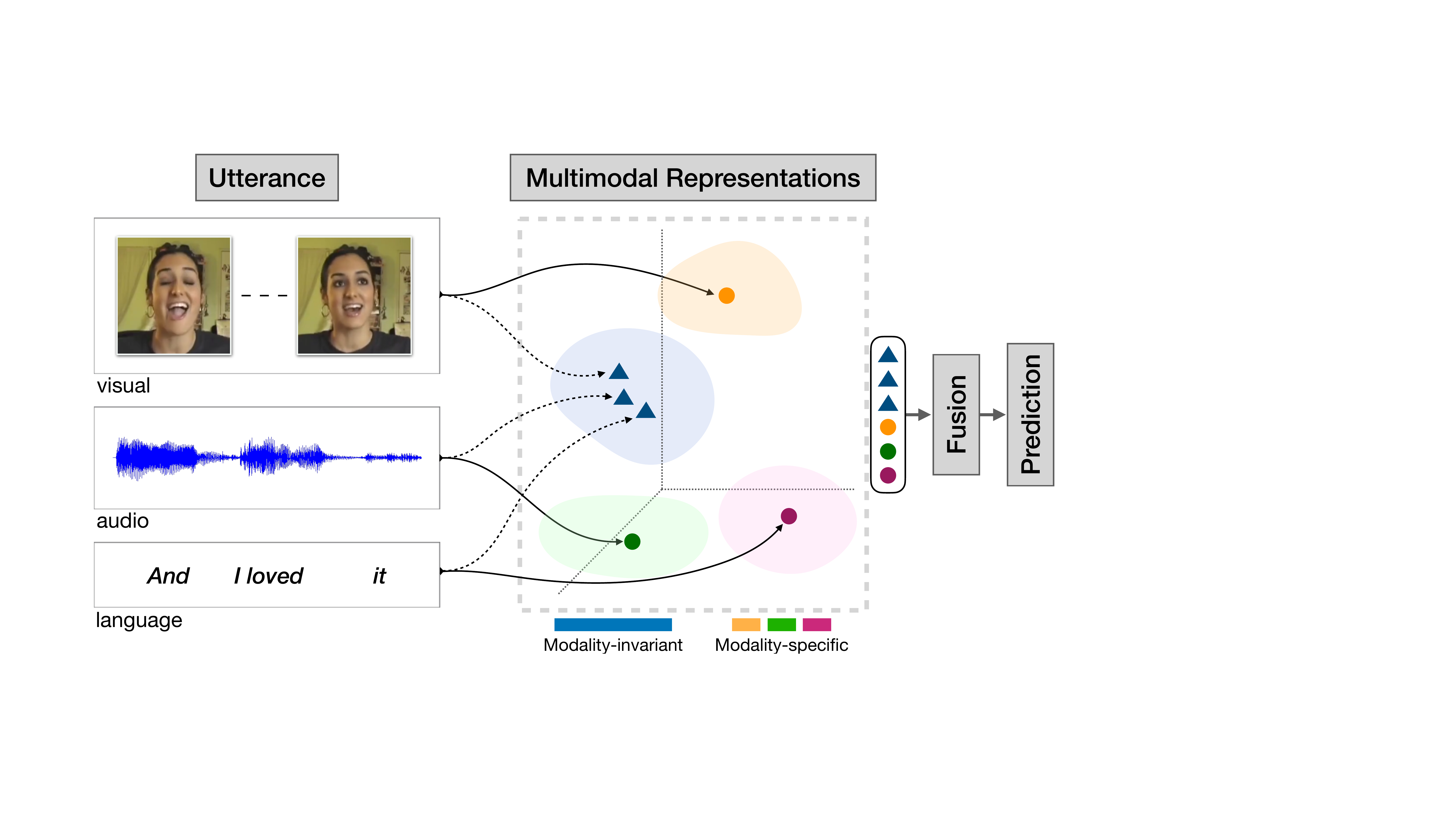}
    \caption{Learning multimodal representations through modality-invariant and -specific subspaces. These features are later utilized for fusion and subsequent prediction of affect in the video.}
    \label{fig:example}
\end{figure}

Motivated by recent advances in domain adaptation~\cite{DBLP:conf/nips/BousmalisTSKE16}, MISA learns two distinct utterance representations for each modality. The first representation is \textit{modality-invariant} and aimed towards reducing modality gaps. Here, all the modalities for an utterance are mapped to a shared subspace with distributional alignment. Though multimodal signals come from different sources, they share common motives and goals of the speaker, which is responsible for the overall affective state of the utterance. The invariant mappings help capture these underlying commonalities and correlated features as aligned projections on the shared subspace. Most of the prior works do not utilize such alignment before fusion, which puts an extra burden on their fusion to bridge modality gaps and learn common features. 

In addition to the invariant subspace, MISA also learns \textit{modality-specific} features that are private to each modality. For any utterance, each modality holds distinctive characteristics that include speaker-sensitive stylistic information. Such idiosyncratic details are often uncorrelated to other modalities and are categorized as noise. Nevertheless, they could be useful in predicting the affective state -- for example, a speaker's tendency to be sarcastic or peculiar expressions biased towards an affective polarity. Learning such modality-specific features, thus, complements the common latent features captured in the invariant space and provides a comprehensive multimodal representation of the utterance. We propose to use this full set of representations for fusion (see~\cref{fig:example}).

To learn these subspaces, we incorporate a combination of losses that include distributional similarity loss (for invariant features), orthogonal loss (for specific features), reconstruction loss (for representativeness of the modality features), and the task prediction loss. We evaluate our hypothesis on two popular benchmark datasets of MSA -- MOSI and MOSEI. We also check the adaptability of our model to another similar task -- \textit{Multimodal Humor Detection} (MHD), where we evaluate the recently proposed UR$\_$FUNNY dataset. In all three cases, we observe strong gains that surpass state-of-the-art models, highlighting the efficacy of MISA.

The novel contributions of this paper can be summarized as:
\begin{itemize}[leftmargin=*]
    \item We propose \textit{MISA} -- a simple and flexible multimodal learning framework that emphasizes on multimodal representation learning as a pre-cursor to multimodal fusion. MISA learns modality-invariant and modality-specific representations to give a comprehensive and disentangled view of the multimodal data, thus aiding fusion for predicting affective states.
    \item Experiments on MSA and MHD tasks demonstrate the power of MISA where the learned representations help a simple fusion strategy surpass complex state-of-the-art models.
\end{itemize}


\section{Related Works} \label{sec:related_works}


\subsection{Multimodal Sentiment Analysis.} \label{sec:related_works_msa}

The literature in MSA can be broadly classified into: $(i)$ \textit{Utterance-level} $(ii)$ \textit{Inter-utterance contextual} models. While utterance-level algorithms consider a target utterance in isolation, contextual algorithms utilize neighboring utterances from the overall video. 

\paragraph{\textbf{Utterance-level.}}
Proposed works in this category have primarily focused on learning cross-modal dynamics using sophisticated fusion mechanisms. These works include variety of methods, such as, multiple kernel learning~\cite{DBLP:conf/emnlp/PoriaCG15}, and tensor-based fusion (including its low-rank variants)~\cite{DBLP:conf/emnlp/ZadehCPCM17,DBLP:conf/emnlp/FukuiPYRDR16,DBLP:conf/iccv/HuHYZLMHRY17, DBLP:conf/acl/MorencyLZLSL18,DBLP:conf/acl/MaiHX19,DBLP:journals/tmm/MaiXH20}. While these works perform fusion over representations of utterances, another line of work takes a fine-grained view to perform fusion at the word level. Approaches include multimodal-aware word embeddings~\cite{DBLP:conf/aaai/WangSLLZM19}, recurrent multi-stage fusion~\cite{DBLP:conf/emnlp/LiangLZM18}, graph-based fusion~\cite{DBLP:conf/acl/MorencyCPLZ18,DBLP:journals/corr/abs-1911-07848}, recurrent networks (RNNs), attention-models, memory mechanisms, and transformer-based models~\cite{DBLP:conf/eccv/RajagopalanMBG16, DBLP:conf/aaai/WangSLLZM19, DBLP:conf/aaai/ZadehLMPCM18, DBLP:conf/aaai/ZadehLPVCM18, DBLP:conf/acl/MorencyCPLZ18, DBLP:conf/eccv/RajagopalanMBG16, DBLP:conf/icmi/ChenWLBZM17, DBLP:conf/acl/TsaiBLKMS19}.

\paragraph{\textbf{Inter-utterance context.}}
These models utilize the context from surrounding utterances of the target utterance. Designed as hierarchical networks, they model individual utterances at the lower level and inter-utterance sequential information in the second level.~\citeauthor{DBLP:conf/acl/PoriaCHMZM17} proposed one of the first models, \textit{bc-LSTM}, which utilized this design along with bi-directional LSTMs for the inter-utterance representation learning, framing the overall problem as a structured prediction (sequence tagging) task~\cite{DBLP:conf/acl/PoriaCHMZM17}. Later works involved either improving fusion using attention~\cite{DBLP:conf/icdm/PoriaCHMZM17,chen2019complementary,DBLP:conf/mm/GuLHFYCZM18}, hierarchical fusion~\cite{DBLP:journals/kbs/MajumderHGCP18}, or developing better contextual modeling~\cite{DBLP:conf/emnlp/GhosalACPEB18,DBLP:conf/naacl/AkhtarCGPEB19,DBLP:conf/emnlp/ChauhanAEB19,chen2019complementary}.

Our work is fundamentally different from these available works. We do not use contextual information and neither focus on complex fusion mechanisms. Instead, we stress the importance of representation learning before fusion. Nevertheless, MISA is flexible to incorporate these above-mentioned components, if required.

\subsection{Multimodal Representation Learning.} \label{sec:related_works_represenatations}

\paragraph{\textbf{Common subspace representations.}}
Works that attempt to learn cross-modal common subspaces can be broadly categorized into: $(i)$ \textit{Translation}-based models which translates one modality to another using methods such as sequence-to-sequence~\cite{pham2018seq2seq2sentiment}, cyclic translations~\cite{DBLP:conf/aaai/PhamLMMP19}, and adversarial auto-encoders~\cite{DBLP:journals/corr/abs-1911-07848}; $(ii)$ Correlation-based models~\cite{DBLP:journals/corr/abs-1911-05544} that learn cross-modal correlations using Canonical Correlation Analysis~\cite{DBLP:conf/icml/AndrewABL13}; $(iii)$ Learning a new shared subspace where all the modalities are simultaneously mapped, using techniques such as adversarial learning~\cite{DBLP:journals/tomccap/PengQ19,DBLP:journals/corr/ParkI16}.  Similar to the third category, we also learn common modality-invariant subspaces. However, we do not use adversarial discriminators to learn shared mappings. Moreover, we incorporate orthogonal modality-specific representations -- a trait less explored in multimodal learning tasks. 

\paragraph{\textbf{Factorized representations.}}
Within the regime of subspace learning, we turn our focus to factorized representations. While one line of work attempts to learn generative-discriminative factors of the multimodal data~\cite{DBLP:conf/iclr/TsaiLZMS19}, our focus is to learn modality-invariant and -specific representations. To achieve this, we take motivation from related literature on shared-private representations. 

The origins of shared-private~\cite{DBLP:conf/nips/BousmalisTSKE16} learning can be found in multi-view component analysis~\cite{DBLP:journals/jmlr/SalzmannEUD10}. These early works designed latent variable models (LVMs) with separate shared and private latent variables~\cite{DBLP:conf/cvpr/SongMD12}. \citet{DBLP:journals/corr/WangLL16a} revisited this framework by proposing a probabilistic CCA -- deep variational CCA. Unlike these models, our proposal involves a discriminative deep neural architecture that obviates the need for approximate inference. 

Our framework is closely related to the \textit{Domain Separation Network} (DSN)~\cite{DBLP:conf/nips/BousmalisTSKE16}, which proposed the shared-private model for domain adaptation. DSN has been influential in the development of similar models in areas such as multi-task text classification~\cite{DBLP:conf/acl/LiuQH17}. Although we derive inspiration from DSN, MISA contains critical distinctions: $(i)$ DSN learns factorized representations across instances, whereas MISA learns the representations for modalities within instances (utterances); $(ii)$ Unlike DSN, we use a more-advanced distribution similarity metric -- CMD (see~\cref{sec:loss_def}) over adversarial training or MMD; $(iii)$ We incorporate additional orthogonal losses across modality-specific (private) representations (see~\cref{sec:diff_loss}); $(iv)$ Finally, while DSN uses only shared representations for task predictions, MISA incorporates both invariant and specific representations for fusion followed by task prediction. We posit that availing both the modality representations helps aid fusion by providing a holistic view of the multimodal data.

\begin{figure*}[t]
    \centering
    \includegraphics[width=\linewidth]{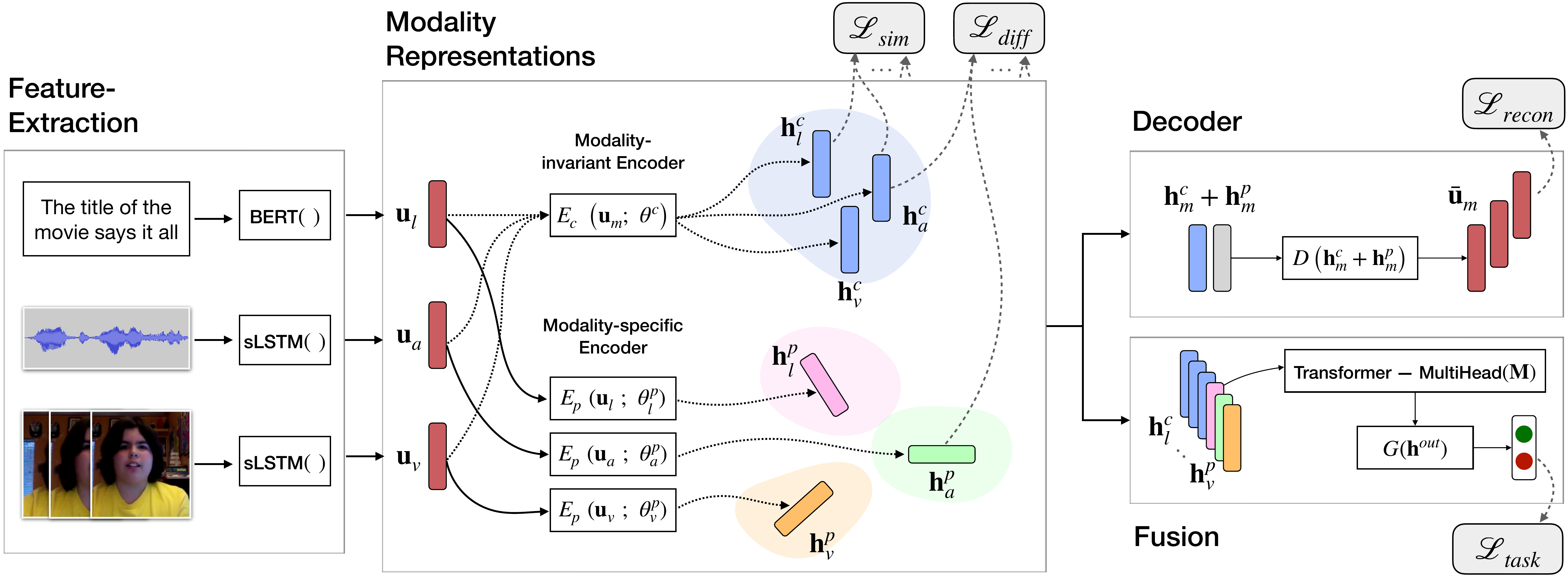}
    \caption{MISA takes the utterance-level representations and projects each modality to two subspaces: modality-invariant and -specific. Later, these hidden representations are used to reconstruct each input and also used for fusion to make the task predictions.}
    \label{fig:framework}
\end{figure*}

\section{Approach} \label{sec:approach}

\subsection{Task Setup} 
Our goal is to detect sentiments in videos by leveraging multimodal signals. Each video in the data is segmented into its constituent utterances~\footnote{An utterance is a unit of speech bounded by breaths or pauses~\cite{olson1977utterance}.}, where each utterance --- a smaller video by itself --- is considered as an input to the model. For an utterance $U$, the input comprises of three sequences of low-level features from language $(l)$, visual $(v)$ and acoustic $(a)$ modalities. These are represented as $\mathbf{U}_l \in \mathbb{R}^{T_{l} \times d_{l}}$, $\mathbf{U}_v \in \mathbb{R}^{T_{v} \times d_{v}}$, and $\mathbf{U}_a \in \mathbb{R}^{T_{a} \times d_{a}}$ respectively. Here $T_{m}$ denotes the length of the utterance, such as number of tokens ($T_{l}$), for modality $m$ and $d_{m}$ denotes the respective feature dimensions. The details of these features are discussed in~\cref{sec:features}. 

Given these sequences $\mathbf{U}_{m \in \{l,v,a\}}$, the primary task is to predict the affective orientation of utterance $U$ from either a predefined set of $C$ categories $y \in \mathbb{R}^{C}$ or as a continuous intensity variable $y \in \mathbb{R}$.

\subsection{MISA}
The functioning of MISA can be segmented into two main stages: Modality Representation Learning (\cref{sec:rep_learn}) and Modality Fusion (\cref{sec:fusion}). The full framework is illustrated in~\cref{fig:framework}.

\subsection{Modality Representation Learning}
\label{sec:rep_learn}

\paragraph{\textbf{Utterance-level Representations}.} Firstly, for each modality $m \in \{l,v,a\}$, we map its utterance sequence $\mathbf{U}_m \in \mathbb{R}^{T_{m} \times d_{m}}$ to a fixed-sized vector $\mathbf{u}_m \in \mathbb{R}^{d_{h}}$. We use a stacked bi-directional Long Short-Term Memory (LSTM)~\cite{DBLP:journals/neco/HochreiterS97} whose end-state hidden representations coupled with a fully connected dense layer gives $\mathbf{u}_m$:
\begin{align}
    \mathbf{u}_m = \text{sLSTM}\left( \ \mathbf{U}_m; \; \theta^{lstm}_m \ \right) \label{eq:feature_extract}
\end{align}

\paragraph{\textbf{Modality-Invariant and -Specific Representations.}}
We now project each of the utterance vector $\mathbf{u}_{m}$ to two distinct representations. First is the modality-invariant component that learns a shared representation in a common subspace with distributional similarity constraints~\cite{DBLP:journals/access/GuoWW19}. This constraint aids in minimizing the heterogeneity gap -- a desirable property for multimodal fusion. Second is the modality-specific component that captures the unique characteristics of that modality. Through this paper, we argue that the presence of both modality-invariant and -specific representations provides a holistic view that is required for effective fusion. Learning these representations is the primary goal of our work.

Given the utterance vector $\mathbf{u}_m$ for modality $m$, we learn the hidden modality-invariant ($\mathbf{h}_{m}^{c}  \in \mathbb{R}^{d_{h}} $ ) and  modality-specific ($\mathbf{h}_{m}^{p}  \in \mathbb{R}^{d_{h}}$) representations using the encoding functions:
\begin{align}
    \mathbf{h}_{m}^{c} = E_{c}\left(\mathbf{u}_m ; \theta^{c} \right) &\; ,\quad \mathbf{h}_{m}^{p} = E_{p}\left(\mathbf{u}_m ; \theta^{p}_{m} \right) \label{eq:modality_reps}
\end{align}
To generate the six hidden vectors $\mathbf{h}_{l/v/a}^{p/c}$ (two per modality), we use simple feed-forward neural layers; $E_c$ shares the parameters $\theta^c$ across all three modalities, whereas $E_p$ assigns separate parameters $\theta_m^p$ for each modality.

\subsection{Modality Fusion}
\label{sec:fusion}

After projecting the modalities into their respective representations, we fuse them into a joint vector for downstream predictions. We design a simple fusion mechanism that first performs a self-attention --- based on the Transformer~\cite{DBLP:conf/nips/VaswaniSPUJGKP17} --- followed by a concatenation of all the six transformed modality vectors.

\begin{remarkTransformer}
The Transformer leverages an attention module that is defined as a scaled dot-product function:
\begin{align}
    \text{Attention(Q, K, V)} = \text{softmax}\left(\frac{\mathbf{Q}\mathbf{K}^T}{\sqrt{d_h}}\right)\mathbf{V}
\end{align}
Where, $\mathbf{Q}$, $\mathbf{K}$, and $\mathbf{V}$ are the \textit{query}, \textit{key}, and \textit{value} matrices. The Transformer computes multiple such parallel attentions, where each attention output is called a \textit{head}. The $i^{th}$ head is computed as:
\begin{align}
    \text{head}_i = \text{Attention}( \ \mathbf{Q}W^q_i, \mathbf{K}W^k_i, \mathbf{V}W^v_i \ ) \label{eq:head_attention}
\end{align}
$W^{q/k/v}_i \in \mathbb{R}^{d_h \times d_h}$ are head-specific parameters to linearly project the matrices into local spaces.
\end{remarkTransformer}

\paragraph{\textbf{Fusion Procedure.}} First we stack the six modality representations (from \cref{eq:modality_reps}) into a matrix $\mathbf{M} = [\mathbf{h}^c_l, \mathbf{h}^c_v, \mathbf{h}^c_a, \mathbf{h}^p_l, \mathbf{h}^p_v, \mathbf{h}^p_a] \in \mathbb{R}^{6 \times d_h}$. Then, we perform a multi-headed self-attention on these representations to make each vector \textit{aware} of the fellow cross-modal (and cross subspace) representations. Doing this allows each representation to induce potential information from fellow representations that are synergistic towards the overall affective orientation. Such cross-modality matching has been highly prominent in recent cross-modal learning approaches~\cite{DBLP:conf/nips/LuBPL19,xi2020multimodal,DBLP:conf/nips/KielaBFT19,DBLP:journals/corr/abs-1908-03557,DBLP:journals/corr/abs-1908-08530}.

For self-attention, we set $\mathbf{Q} = \mathbf{K} = \mathbf{V} = \mathbf{M} \in \mathbb{R}^{6 \times d_h}$.  The Transformer generates a new matrix $\mathbf{\bar{M}} = [\mathbf{\bar{h}}^c_l, \mathbf{\bar{h}}^c_v, \mathbf{\bar{h}}^c_a, \mathbf{\bar{h}}^p_l, \mathbf{\bar{h}}^p_v, \mathbf{\bar{h}}^p_a]$ as:
\begin{align}
    \mathbf{\bar{M}} = \text{MultiHead}(\mathbf{M}; \theta^{att}) = (\text{head$_1$} \oplus \dots \oplus \text{head$_n$})W^o \label{eq:multihead}
\end{align}
where, each head$_i$ here is calculated based on~\cref{eq:head_attention}; $\oplus$ represents concatenation; and $\theta^{att} = \{W^q, W^k, W^v, W^o\}$.

\paragraph{\textbf{Prediction/Inference.}} Finally, we take the Transformer output and construct a joint-vector using concatenation, $\mathbf{h}^{out} = [\mathbf{\bar{h}}^c_l \oplus \dots \oplus \mathbf{\bar{h}}^p_a] \in \mathbb{R}^{6d_h}$. The task predictions are then generated by the function $\mathbf{\hat{y}} = G(\mathbf{h}^{out}; \theta^{out})$.

We provide the network topology of the functions $sLSTM()$, $E_c()$, $E_p()$, $G()$ and $D()$ (explained later) in the appendix.

\subsection{Learning} \label{sec:loss_def}

The overall learning of the model is performed by minimizing:
\begin{align}
    \mathcal{L} =  \mathcal{L}_{\text{task}} + \alpha \,  \mathcal{L}_{\text{sim}}  + \beta \, \mathcal{L}_{\text{diff}} + \gamma\, \mathcal{L}_{\text{recon}} 
\end{align}

Here, $\alpha, \beta, \gamma$ are the interaction weights that determine the contribution of each regularization component to the overall loss $\mathcal{L}$. Each of these component losses are responsible for achieving the desired subspace properties. We discuss them next.

\subsubsection{\textbf{$\mathcal{L}_{\text{sim}}$ -- Similarity Loss}} \label{sec:sim_loss}

Minimizing the \textit{similarity loss} reduces the discrepancy between the shared representations of each modality. This helps the common cross-modal features to be aligned together in the shared subspace. Amongst many choices, we use the \textit{Central Moment Discrepancy} (CMD)~\cite{DBLP:conf/iclr/ZellingerGLNS17} metric for this purpose. CMD is a state-of-the-art distance metric that measures the discrepancy between the distribution of two representations by matching their order-wise moment differences. Intuitively, CMD distance decreases as two distributions become more similar.

\begin{remarkCMD}
Let $X$ and $Y$ be bounded random samples with respective probability distributions $p$ and $q$ on the interval $[a, b]^{N}$. The central moment discrepancy regularizer CMD $_{K}$ is defined as an empirical estimate of the CMD metric, by
\begin{align}
CMD_{K}(X, Y) &= \frac{1}{|b-a|}\|\mathbf{E}(X)-\mathbf{E}(Y)\|_{2} \nonumber \\ 
& + \sum_{k=2}^{K} \frac{1}{|b-a|^{k}}\left\|C_{k}(X)-C_{k}(Y)\right\|_{2}
\end{align}

where, $\mathbf{E}(X)=\frac{1}{|X|} \sum_{x \in X} x$ is the empirical expectation vector of sample $X$ and $C_{k}(X)=\mathbf{E}\left((x-\mathbf{E}(X))^{k}\right)$ is the vector of all $k^{t h}$ order sample central moments of the coordinates of $X$. 
\end{remarkCMD}

In our case, we calculate the CMD loss between the invariant representations of each pair of modalities:
\begin{align}
    \mathcal{L}_{\text{sim}} = \frac{1}{3} \sum_{\substack{(m_1,m_2) \in \\ \{(l,a),(l,v), \\ (a,v)\}}}{ CMD_K(\mathbf{h}_{m_1}^{c}, \mathbf{h}_{m_2}^{c}) }  \label{eq:cmd_loss}
\end{align}

Here, we make two important observations: $(i)$ We choose CMD over KL-divergence or MMD because CMD is a popular metric~\cite{DBLP:conf/acl/ZhangHPJ18} and performs explicit matching of higher-order moments without expensive distance and kernel matrix computations. $(ii)$ \textit{Adversarial loss} is another choice for similarity training, where a discriminator and the shared encoder engage in a minimax game. However, we choose CMD owing to its simple formulation. In contrast, adversarial training demands additional parameters for the discriminator along with added complexities, such as oscillations in training~\cite{DBLP:series/acvpr/HoffmanTDS17}.

\subsubsection{\textbf{$\mathcal{L}_{\text{diff}}$ -- Difference Loss}} \label{sec:diff_loss}

This loss is to ensure that the modality-invariant and -specific representations capture different aspects of the input. The non-redundancy is achieved by enforcing a soft orthogonality constraint between the two representations~\cite{DBLP:conf/nips/BousmalisTSKE16,DBLP:conf/acl/PlankR18,DBLP:conf/acl/LiuQH17}. In a training batch of utterances, let $\mathbf{H}_m^c$ and $\mathbf{H}_m^p$ be the matrices~\footnote{We transform the matrices to have zero mean and unit $l_2$ norm.} whose rows denote the hidden vectors $\mathbf{h}_m^{c}$ and $\mathbf{h}_m^{p}$ for modality $m$ of each utterance. Then the orthogonality constraint for this modality vector pair is calculated as:
\begin{align}
    \left\|\mathbf{H}_{m}^{c^{\top}} \mathbf{H}_{m}^{p}\right\|_{F}^{2}
\end{align}
Here, $\|\cdot\|_{F}^{2}$ is the squared Frobenius norm. In addition to the constraints between the invariant and specific vectors, we also add orthogonality constraints between the modality-specific vectors. The overall difference loss is then computed as:
\begin{align}
    \mathcal{L}_{\text{diff}} &=  \sum_{m \in \{l,v,a\}}{\left\|\mathbf{H}_{m}^{c^{\top}} \mathbf{H}_{m}^{p}\right\|_{F}^{2}} + \sum_{ \substack{(m_1,m_2) \in \\ \{(l,a),(l,v), \\ (a,v)\}}}{\left\|\mathbf{H}_{m_1}^{p^{\top}} \mathbf{H}_{m_2}^{p}\right\|_{F}^{2}} 
\end{align}

\subsubsection{\textbf{$\mathcal{L}_{\text{recon}}$ -- Reconstruction Loss}} \label{sec:recon_loss}
As the \textit{difference loss} is enforced, there remains a risk of learning trivial representations by the modality-specific encoders.  Trivial cases can arise if the encoder function approximates an orthogonal but unrepresentative vector of the modality. To avoid this situation, we add a \textit{reconstruction loss} that ensures the hidden representations to capture details of their respective modality. First, we reconstruct the modality vector $\mathbf{u}_m$ by using a decoder function $\mathbf{\hat{u}}_m = D( \mathbf{h}_{m}^{c} + \mathbf{h}_{m}^{p}; \theta^{d})$. The reconstruction loss is then the \textit{mean squared error} loss between $\mathbf{u}_m$ and $\mathbf{\hat{u}}_m$:
\begin{align}
    \mathcal{L}_{\text{recon}} &= \frac{1}{3} \left( \sum_{m \in \{l,v,a\}}{ \frac{\|\mathbf{u}_m-\mathbf{\hat{u}}_m\|_{2}^{2}}{d_h}  }\right)
\end{align}
Where, $\|\cdot\|_{2}^{2}$ is the squared $L_2$-norm.

\subsubsection{\textbf{$\mathcal{L}_{\text{task}}$ -- Task Loss}} \label{sec:task_loss}
The task-specific loss estimates the quality of prediction during training. For classification tasks, we use the standard \textit{cross-entropy loss} whereas for regression tasks, we use \textit{mean squared error} loss. For $N_b$ utterances in a batch, these are calculated as:
\begin{align}
    \mathcal{L}_{\text {task }} &=- \frac{1}{N_b} \sum_{i=0}^{N_{b}} \mathbf{y}_{i} \cdot \log \hat{\mathbf{y}}_{i} && \text{for classification}\\
    &= \frac{1}{N_b} \sum_{i=0}^{N_{b}} \|\mathbf{y}_{i} - \hat{\mathbf{y}}_{i}\|_2^2 && \text{for regression}
\end{align}

\section{Experiments} \label{sec:experiments}

\subsection{Datasets} \label{sec:datasets}

We consider benchmark datasets for both the tasks of MSA and MHD. These datasets provide word-aligned multimodal signals (language, visual, and acoustic) for each utterance. 

\subsubsection{\textbf{CMU-MOSI}} 
The CMU-MOSI dataset~\cite{DBLP:journals/expert/ZadehZPM16} is a popular benchmark dataset for research in MSA. The dataset is a collection of YouTube monologues, where a speaker expresses their opinions on topics such as movies. With a total of 93 videos, spanning 89 distance speakers, MOSI contains $2198$ subjective utterance-video segments. The utterances are manually annotated with a continuous opinion score between $[-3,3]$, where $-3/+3$ represents strongly negative/positive sentiments.

\subsubsection{\textbf{CMU-MOSEI}} 
The CMU-MOSEI dataset~\cite{DBLP:conf/acl/MorencyCPLZ18} is an improvement over MOSI with higher number of utterances, greater variety in samples, speakers, and topics. The dataset contains $23453$ annotated video segments (utterances), from $5000$ videos, $1000$ distinct speakers and $250$ different topics. 

\subsubsection{\textbf{UR$\_$FUNNY}} 
For MHD, we consider the recently proposed UR$\_$FUNNY dataset~\cite{DBLP:conf/emnlp/HasanRZZTMH19}. Similar to sentiments, generating and perceiving humor also occurs through multimodal channels. This dataset, thus provides multimodal utterances that act as punchlines sampled from TED talks. It also provides associated context for each target utterance and ensures diversity in both speakers and topics. Each target utterance is labeled with a binary label for humor/non-humor instance. Dataset split and training details are available in the appendix.

\subsection{Evaluation Criteria} \label{sec:metrics}

Sentiment intensity prediction in both MOSI and MOSEI datasets are regression tasks with \textit{mean absolute error} (MAE) and \textit{Pearson correlation} (Corr) as the metrics. Additionally, the benchmark also involves classification scores that include, \textit{seven-class accuracy (Acc-7)} ranging from $-3$ to $3$, \textit{binary accuracy (Acc-2)} and \textit{F-Score}. For binary accuracy scores, two distinct approaches have been considered in the past. First is \textit{negative}/\textit{non-negative} classification where the labels for non-negatives are based on scores being $\geq 0$~\cite{DBLP:conf/aaai/ZadehLPVCM18}. In recent works, binary accuracy is calculated on the more accurate formulation of \textit{negative}/\textit{positive} classes where negative and positive classes are assigned for $< 0$ and $> 0$ sentiment scores, respectively~\cite{DBLP:conf/acl/TsaiBLKMS19}. We report results on both these metrics using the segmentation marker $- / -$ where the left-side score is for \textit{neg.}/\textit{non-neg.} while the right-side score is for \textit{neg.}/\textit{pos.} classification. For UR$\_$FUNNY dataset, the task is a standard binary classification with binary accuracy (Acc-2) as the metric for evaluation~\cite{DBLP:conf/emnlp/HasanRZZTMH19}.

\subsection{Feature Extraction} \label{sec:features}

For fair comparisons, we utilize the standard low-level features that are provided by the respective benchmarks and utilized by the state-of-the-art methods.

\subsubsection{\textbf{Language Features}}
Traditionally, language modality features has been GloVe~\cite{DBLP:conf/emnlp/PenningtonSM14} embeddings for each token in the utterance. However, following recent works~\cite{chen2019complementary}, including the state-of-the-art ICCN~\cite{DBLP:journals/corr/abs-1911-05544}, we utilize the pre-trained \verb|BERT|~\cite{DBLP:conf/naacl/DevlinCLT19} as the feature extractor for textual utterances. Using \verb|BERT| replaces the $sLSTM(\mathbf{U}_l; \theta^{lstm}_l)$ in~\cref{eq:feature_extract} with \verb|BERT|($U_l; \theta^{bert}$). For UR$\_$FUNNY, however, the state of the art is based on GloVe features. Thus, for fair comparison, we provide results using both GloVe and \verb|BERT|.

While GloVe features are $300$ dimensional token embeddings, for \verb|BERT|, we utilize the \textit{BERT-base-uncased} pre-trained model. This model comprises of $12$ stacked Transformer layers. Aligned with recent works~\cite{DBLP:journals/corr/abs-2004-02105}, we choose the utterance vector $\mathbf{u}_l$ to be the average representation of the tokens from the final $768$ dimensional hidden state. Unfortunately, for our considered UR$\_$FUNNY version, the original transcripts are not available. Instead, only the GloVe embeddings have been provided. To retrieve the raw texts, we choose the token with the least cosine distance from the GloVe vocabulary for each word embedding. A manual check of $100$ randomly sampled utterances validated the quality of this process to retrieve legible original transcripts.

\subsubsection{\textbf{Visual Features}}

Both MOSI and MOSEI use Facet~\footnote{\url{https://imotions.com/platform/}} to extract facial expression features, which include facial action units and face pose based on the Facial Action Coding System (FACS)~\cite{ekman1997face}. This process is repeated for each sampled frame within the utterance video sequence. For UR$\_$FUNNY, OpenFace~\cite{DBLP:conf/wacv/Baltrusaitis0M16}, a facial behavioral analysis tool, is used to extract features related to the facial expressions of the speaker. The final visual feature dimensions, $d_{v}$, are $47$ for MOSI, $35$ for MOSEI, and $75$ for UR$\_$FUNNY.

\subsubsection{\textbf{Acoustic Features}}

The acoustic features contain various low-level statistical audio functions extracted from COVAREP~\cite{DBLP:conf/icassp/DegottexKDRS14} -- an acoustic analysis framework. Some of the features include 12 Mel-frequency ceptral coefficients, pitch, Voiced/Unvoiced segmenting features (VUV)~\cite{DBLP:conf/interspeech/DrugmanA11}, glottal source parameters~\cite{DBLP:journals/taslp/DrugmanTGND12}, and other features related to emotions and tone of speech~\footnote{Please refer to ~\cite{DBLP:conf/acl/MorencyCPLZ18,DBLP:conf/emnlp/HasanRZZTMH19} and their respective SDKs (\url{https://github.com/A2Zadeh/CMU-MultimodalSDK} v1.1.1; \url{https://github.com/ROC-HCI/UR-FUNNY/blob/master/UR-FUNNY-V1.md} v1) for a full list of the features. Sampling rates for the acoustic and visual signals are summarized in~\cite{DBLP:conf/acl/TsaiBLKMS19}. Following related works, we align all three modalities based on language modality. This standard procedure makes all three temporal sequences within an utterance to be of equal length, i.e. $T_l = T_v = T_a$}. The feature dimensions, $d_{a}$, are $74$ for MOSI/MOSEI and $81$ for UR$\_$FUNNY.

\subsection{Baselines} \label{sec:baselines}

We perform a comprehensive comparative study against MISA by considering various baselines as detailed below.

\subsubsection{\textbf{Previous Models.}}

Numerous methods have been proposed for multimodal learning, especially for sentiment analysis and human language tasks in general. As mentioned in~\cref{sec:related_works}, these works can be broadly categorized into utterance-level and inter-utterance contextual models. Utterance-level baselines include:

\begin{itemize}[leftmargin=*]
    \item Networks which perform temporal modeling and fusion of utterances: \textbf{MFN}~\cite{DBLP:conf/aaai/ZadehLMPCM18}, \textbf{MARN}~\cite{DBLP:conf/aaai/ZadehLPVCM18}, \textbf{MV-LSTM}~\cite{DBLP:conf/eccv/RajagopalanMBG16}, \textbf{RMFN}~\cite{DBLP:conf/emnlp/LiangLZM18}.
    \item Models which utilize attention and transformer modules to improve token representations using non-verbal signals: \textbf{RAVEN}~\cite{DBLP:conf/aaai/WangSLLZM19}, \textbf{MulT}~\cite{DBLP:conf/acl/TsaiBLKMS19}.
    \item Graph-based fusion models: \textbf{Graph-MFN}~\cite{DBLP:conf/acl/MorencyCPLZ18}.
    \item Utterance-vector fusion approaches that use tensor-based fusion and low-rank variants: \textbf{TFN}~\cite{DBLP:conf/emnlp/ZadehCPCM17}, \textbf{LMF}~\cite{DBLP:conf/acl/MorencyLZLSL18}, \textbf{LMFN}~\cite{DBLP:journals/tmm/MaiXH20}, \textbf{HFFN}~\cite{DBLP:conf/acl/MaiHX19}.
    \item Common subspace learning models that use cyclic translations (\textbf{MCTN}~\cite{DBLP:conf/aaai/PhamLMMP19}), adversarial auto-encoders (\textbf{ARGF}~\cite{DBLP:journals/corr/abs-1911-07848}), and generative-discriminative factorized representations (\textbf{MFM}~\cite{DBLP:conf/iclr/TsaiLZMS19}).
\end{itemize}

\noindent Inter-utterance contextual baselines include:

\begin{itemize}[leftmargin=*]
    \item RNN-based models: \textbf{BC-LSTM}~\cite{DBLP:conf/acl/PoriaCHMZM17}, with hierarchical fusion -- \textbf{CH-Fusion}~\cite{DBLP:journals/kbs/MajumderHGCP18}.
    \item Inter-utterance attention and multi-tasking models: \textbf{CIA}~\cite{DBLP:conf/emnlp/ChauhanAEB19},  \textbf{CIM-MTL}~\cite{DBLP:conf/naacl/AkhtarCGPEB19}, \textbf{DFF-ATMF}~\cite{chen2019complementary}.
\end{itemize}
For detailed descriptions of the models, please refer to the appendix.

\subsubsection{\textbf{State of the Art.}} 

For the task of MSA, the Interaction Canonical Correlation Network (\textbf{ICCN})~\cite{DBLP:journals/corr/abs-1911-05544} stands as the state-of-the-art (SOTA) model on both MOSI and MOSEI. ICCN first extracts features from audio and video modality and then fuses with text embeddings to get two outer products, text-audio and text-video. Finally, the outer products are fed to a Canonical Correlation Analysis (CCA) network, whose output is used for prediction.

For MHD, The SOTA is Contextual Memory Fusion Network (\textbf{C-MFN})~\cite{DBLP:conf/emnlp/HasanRZZTMH19}, which extends the MFN model by proposing uni- and multimodal context networks that consider preceding utterances and performs fusion using the MFN model as its backbone. Originally,  MFN~\cite{DBLP:conf/aaai/ZadehLMPCM18} is a multi-view gated memory network that stores intra- and cross-modal utterance interactions in its memories.

\section{Results and Analysis} \label{sec:results}

\begin{table}[t]
  \centering
  \resizebox{\linewidth}{!}{
  \begin{tabular}{|l|c:c:c:c:c|}
    \hline
    \multirow{2}{*}{Models} & \multicolumn{5}{c|}{MOSI}\\
     & \multicolumn{1}{c}{MAE ($\downarrow$)} & \multicolumn{1}{c}{Corr ($\uparrow$)} & \multicolumn{1}{c}{Acc-2 ($\uparrow$)} & \multicolumn{1}{c}{F-Score ($\uparrow$)} & \multicolumn{1}{c|}{Acc-7 ($\uparrow$)}  \\
    \hline
    \hline
    BC-LSTM & 1.079 & 0.581 & 73.9 / - & 73.9 / - & 28.7 \\
    MV-LSTM & 1.019 & 0.601 & 73.9 / - & 74.0 / - & 33.2 \\ 
    TFN &  0.970 & 0.633 & 73.9 / - & 73.4 / - & 32.1 \\
    MARN &  0.968 & 0.625 & 77.1 / - & 77.0 / - & 34.7 \\
    MFN & 0.965 & 0.632 & 77.4 / - & 77.3 / - & 34.1 \\
    LMF & 0.912 & 0.668 & 76.4 / - & 75.7 / - & 32.8 \\
    CH-Fusion & - & - & 80.0 / - & -  & - \\
    MFM$^{\otimes}$ & 0.951 & 0.662 & 78.1 / - & 78.1 / - & 36.2 \\
    RAVEN$^{\otimes}$ & 0.915 & 0.691 & 78.0 / - & 76.6 / - & 33.2 \\
    RMFN$^{\otimes}$ & 0.922 & 0.681 & 78.4 / - & 78.0 / - & 38.3 \\
    MCTN$^{\otimes}$ & 0.909 & 0.676 & 79.3 / - & 79.1 / - & 35.6 \\
    CIA& 0.914 & 0.689 & 79.8 / - & - / 79.5 & 38.9\\
    HFFN$^{\oslash}$ & - & - & - / 80.2 & - / 80.3 & - \\
    LMFN$^{\oslash}$ & - & - & - / 80.9 & - / 80.9 & - \\
    DFF-ATMF (B) & - & - & - / 80.9 & - / 81.2 & -\\
    ARGF & - & - & - / 81.4 & - / 81.5 & - \\
    MulT & 0.871 & 0.698 & - / 83.0 & - / 82.8 & 40.0 \\
    TFN (B)$^{\diamond}$ & 0.901 & 0.698 & - / 80.8  & - / 80.7 & 34.9 \\
    LMF (B)$^{\diamond}$ & 0.917 & 0.695 & - / 82.5  & - / 82.4 & 33.2 \\
    MFM (B)$^{\diamond}$ & 0.877 & 0.706 & - / 81.7 & - / 81.6 & 35.4 \\
    ICCN (B) & 0.860  & 0.710 & - / 83.0  & - / 83.0 & 39.0 \\
    \hline
    MISA (B)& \textbf{0.783} & \textbf{0.761} & \textbf{81.8}$^{\dagger}$ / \textbf{83.4}$^{\dagger}$ & \textbf{81.7} / \textbf{83.6} & \textbf{42.3}\\
    $\Delta_{SOTA}$ & \textcolor{darkgreen}{\(\downarrow 0.077 \)}  & \textcolor{darkgreen}{\(\uparrow 0.051 \)}  & \textcolor{darkgreen}{\(\uparrow 2.0 \)} / \textcolor{darkgreen}{\(\uparrow 0.4 \)} &  \textcolor{darkgreen}{\(\uparrow 2.6 \)} / \textcolor{darkgreen}{\(\uparrow 0.6 \)} &   \textcolor{darkgreen}{\(\uparrow 3.3 \)}\\
    \hline
  \end{tabular}
  }
  \caption{Performances of multimodal models in MOSI. NOTE: (B) means the language features are based on BERT; $^{\otimes}$ from~\cite{DBLP:conf/acl/TsaiBLKMS19}; $^{\oslash}$ from~\cite{DBLP:journals/corr/abs-1911-07848}; $^{\diamond}$ from~\cite{DBLP:journals/corr/abs-1911-05544}. Final row presents our best model per metric. $^{\dagger} p < 0.05$ under McNemar's Test for binary classification. Here, the statistical significance tests are compared with publicly available models of~\cite{DBLP:conf/emnlp/ZadehCPCM17,DBLP:conf/acl/MorencyLZLSL18,DBLP:conf/iclr/TsaiLZMS19}.}
  \label{tab:sota_mosi}
\end{table}

\begin{table}[t]
  \centering
  \resizebox{\linewidth}{!}{
  \begin{tabular}{|l|c:c:c:c:c|}
    \hline
    \multirow{2}{*}{Models} & \multicolumn{5}{c|}{MOSEI}\\
     & \multicolumn{1}{c}{MAE ($\downarrow$)} & \multicolumn{1}{c}{Corr ($\uparrow$)} & \multicolumn{1}{c}{Acc-2 ($\uparrow$)} & \multicolumn{1}{c}{F-Score ($\uparrow$)} & \multicolumn{1}{c|}{Acc-7 ($\uparrow$)}  \\
    \hline
    \hline
    MFN$^{\otimes}$ & - & - & 76.0 / - & 76.0 / - & - \\
    MV-LSTM$^{\otimes}$ & - & - & 76.4 / - & 76.4 / - & - \\
    Graph-MFN$^{\otimes}$  & 0.710 & 0.540 & 76.9 / - & 77.0 / -  &  45.0 \\
    RAVEN  &  0.614 & 0.662 & 79.1 / -  & 79.5 / -  & 50.0\\
    MCTN & 0.609 & 0.670 & 79.8 / -  & 80.6 / -  & 49.6\\
    CIA & 0.680 & 0.590 & 80.4 / -  & 78.2 / - & 50.1 \\
    CIM-MTL & - & - & 80.5 / - & 78.8 / -  & - \\
    DFF-ATMF (B)& - & - & - / 77.1 & - / 78.3 & -\\
    MulT & 0.580 & 0.703 & - / 82.5 & - / 82.3 & 51.8 \\
    TFN (B)$^{\diamond}$ & 0.593 & 0.700 & - / 82.5 & - / 82.1 & 50.2 \\
    LMF (B)$^{\diamond}$ & 0.623 & 0.677 & - / 82.0  & - / 82.1 & 48.0 \\
    MFM (B)$^{\diamond}$ & 0.568 & 0.717 & - / 84.4 & - / 84.3 & 51.3 \\
    ICCN (B)& 0.565 & 0.713 & - / 84.2 & - / 84.2 & 51.6 \\
    \hline
    MISA (B) & \textbf{0.555} & \textbf{0.756} & \textbf{83.6}$^{\dagger}$ / \textbf{85.5}$^{\dagger}$ & \textbf{83.8} / \textbf{85.3} & \textbf{52.2} \\
    $\Delta_{SOTA}$ & \textcolor{darkgreen}{\(\downarrow 0.010 \)}  & \textcolor{darkgreen}{\(\uparrow 0.043 \)}  & \textcolor{darkgreen}{\(\uparrow 3.1 \)} / \textcolor{darkgreen}{\(\uparrow 1.3 \)} &  \textcolor{darkgreen}{\(\uparrow 5.0 \)} / \textcolor{darkgreen}{\(\uparrow 1.1 \)} &   \textcolor{darkgreen}{\(\uparrow 0.6 \)}\\
    \hline
  \end{tabular}
  }
  \caption{Performances of multimodal models in MOSEI. NOTE: (B) means the language features are based on BERT; $^{\otimes}$ from~\cite{DBLP:conf/acl/MorencyCPLZ18}; $^{\diamond}$ from~\cite{DBLP:journals/corr/abs-1911-05544}. Final row presents our best model per metric. $^{\dagger} p < 0.05$ under McNemar's Test for binary classification (compared with publicly available models of~\cite{DBLP:conf/emnlp/ZadehCPCM17,DBLP:conf/acl/MorencyLZLSL18,DBLP:conf/iclr/TsaiLZMS19}).}
  \label{tab:sota_mosei}
\end{table}

\subsection{Quantitative Results}

\subsubsection{\textbf{Multimodal Sentiment Analysis}}
The comparative results for MSA are presented in~\cref{tab:sota_mosi} (MOSI) and~\cref{tab:sota_mosei} (MOSEI). In both the datasets, MISA achieves the best performance and surpasses the baselines --- including the state-of-the-art ICCN --- across all metrics (regression and classification combined). Within the results, it can be seen that our model, which is an utterance-level model, fares better than the contextual models. This is an encouraging result as we are able to perform better even with lesser information. Our model also surpasses some of the intricate fusion mechanisms, such as TFN and LFN, which justify the importance of learning multimodal representations preceding the fusion stage. 

\subsubsection{\textbf{Multimodal Humor Detection}}

Similar trends are observed for MHD (see~\cref{tab:sota_ur_funny}), with a highly pronounced improvement over the contextual SOTA, C-MFN. This is true even while using GloVe features for language modality. In fact, our GloVe variant is at par to the \verb|BERT|-based baselines, such as TFN. This indicates that effective modeling of multimodal representations goes a long way. Humor detection is known to be highly sensitive to the idiosyncratic characteristics of different modalities~\cite{DBLP:conf/emnlp/HasanRZZTMH19}. Such dependencies are well modeled by our representations, which is reflected in the results.

\begin{table}[t]
  \centering
  \resizebox{.7\linewidth}{!}{
  \begin{tabular}{|lcc|c|}
    \hline
    \multirow{2}{*}{Algorithms} & context & target & \multicolumn{1}{c|}{UR$\_$FUNNY}\\
     &&& \multicolumn{1}{c|}{Accuracy-2 ($\uparrow$)} \\
    \hline
    \hline
    C-MFN& \checkmark & & 58.45 \\
    C-MFN& & \checkmark & 64.47 \\
    TFN & & \checkmark & 64.71 \\
    LMF & & \checkmark & 65.16 \\
    C-MFN & \checkmark & \checkmark & 65.23 \\ 
    LMF (Bert) & & \checkmark & 67.53 \\
    TFN (Bert) & & \checkmark & 68.57 \\
    \hline
    MISA (GloVe) & & \checkmark & 68.60 \\
    MISA (Bert) & & \checkmark & \textbf{70.61}$^{\dagger}$ \\
    $\Delta_{SOTA}$ & & &  \textcolor{darkgreen}{\(\uparrow  2.07 \)} \\
    \hline
  \end{tabular}
  }
  \caption{Performances of multimodal models in UR$\_$FUNNY. $^{\dagger} p < 0.05$ under McNemar's Test for binary classification when compared against~\cite{DBLP:conf/emnlp/ZadehCPCM17,DBLP:conf/acl/MorencyLZLSL18}. \textit{Context}-based models use additional data that include the utterances preceding the target punchline.}
  \label{tab:sota_ur_funny}
\end{table}

\subsubsection{\textbf{BERT vs. GloVe.}} 
In our experiments, we observe improvements in performance when using \verb|BERT| over the traditional GloVe-based features for language. This raises the question as to whether our performance improvements are \textit{solely} due to \verb|BERT| features. To find an answer, we look at the state-of-the-art approach ICCN, which is also based on \verb|BERT|. Our model comfortably beats ICCN in all metrics, through which we can infer that the improvements in multimodal modeling are a critical factor.

\begin{table}[t]
  \centering
  \resizebox{\linewidth}{!}{
  \begin{tabular}{|ll|cc|cc|c|}
    \hline
    &Model &\multicolumn{2}{c}{MOSI}&\multicolumn{2}{c}{MOSEI}&\multicolumn{1}{c|}{UR$\_$FUNNY}\\
    &&  \multicolumn{1}{c}{MAE ($\downarrow$)} & \multicolumn{1}{c}{Corr ($\uparrow$)} & \multicolumn{1}{c}{MAE ($\downarrow$)} & \multicolumn{1}{c}{Corr ($\uparrow$)} & \multicolumn{1}{c|}{Acc-2 ($\uparrow$)} \\
    \hline
    \hline
    1) & MISA & \textbf{0.783} & \textbf{0.761} & \textbf{0.555} & \textbf{0.756} & \textbf{70.6} \\
    \hline
    \hline
    2) & (-) language $l$ & 1.450 & 0.041 & 0.801 & 0.090 & 55.5 \\
    3) & (-) visual $v$ & 0.798 & 0.756 & 0.558 & 0.753 & 69.7 \\
    4) & (-) audio $a$ & 0.849 & 0.732 & 0.562 & 0.753 & 70.2 \\
    \hline 
    \hline
    5) & (-) $\mathcal{L}_{\text{sim}}$ & 0.807 & 0.740 & 0.566 & 0.751 & 69.3 \\
    6) & (-) $\mathcal{L}_{\text{diff}}$ & 0.824 & 0.749 & 0.565 & 0.742 & 69.3 \\
    7) & (-) $\mathcal{L}_{\text{recon}}$ & 0.794 & 0.757 & 0.559 & 0.754 & 69.7 \\
    \hline
    \hline
    8) & \textit{base} & 0.810  & 0.750 & 0.568 & 0.752 & 69.2 \\
    9) & \textit{inv} & 0.811 & 0.737 & 0.561 & 0.743 & 68.8 \\
    10) & \textit{sFusion} & 0.858 & 0.716 & 0.563 & 0.752 & 70.1 \\
    11) & \textit{iFusion} & 0.850 & 0.735 & \textbf{0.555} & 0.750 & 69.8 \\
    \hline
  \end{tabular}
  }
  \caption{Ablation Study. Here, $(-)$ represents removal for the mentioned factors. Model 1 represents the best performing model in each dataset; Model 2,3,4 depicts the effect of individual modalities; Model 5,6,7 presents the effect of regularization; Model 8,9,10,11 presents the variants of MISA as defined in~\cref{sec:role_of_subspaces}. }
  \label{tab:ablation}
\end{table}

\subsection{Ablation Study}

\subsubsection{\textbf{Role of Modalities.}}

In~\cref{tab:ablation} (model $2,3,4$) we remove one modality at a time to observe the effect in performance. Firstly, we see that a multimodal combination provides the best performance, indicating that the model can learn complementary features. Without this case, the tri-modal combination would not fare better than bi-modal variants such as language-visual MISA.

Next, we observe that the performance sharply drops when the language modality is removed. Similar drops are not observed in removing the other two modalities, showing that the text modality has significant dominance over the audio and visual modalities. There could be two reasons for this: 1) The data quality of text modality could be inherently better as they are manual transcriptions. In contrast, audio and visual signals are unfiltered raw signals. 2) \verb|BERT| is a pre-trained model with better expressive power over the randomly initialized audio and visual feature extractor, giving better utterance-level features.  These observations, however, are dataset specific and can not be generalized to any multimodal scenario.

\subsubsection{\textbf{Role of Regularization.}}

Regularization plays a critical role in achieving the desired representations discussed in~\cref{sec:loss_def}. In this section, we first observe how well the losses are learned in the model while training and whether the validation sets follow similar trends. Next, we perform qualitative verification by looking at the feature distributions of the learned models. Finally, we look at the importance of each loss by an ablation study.

\begin{figure}[t]
    \centering
    \includegraphics[width=\linewidth]{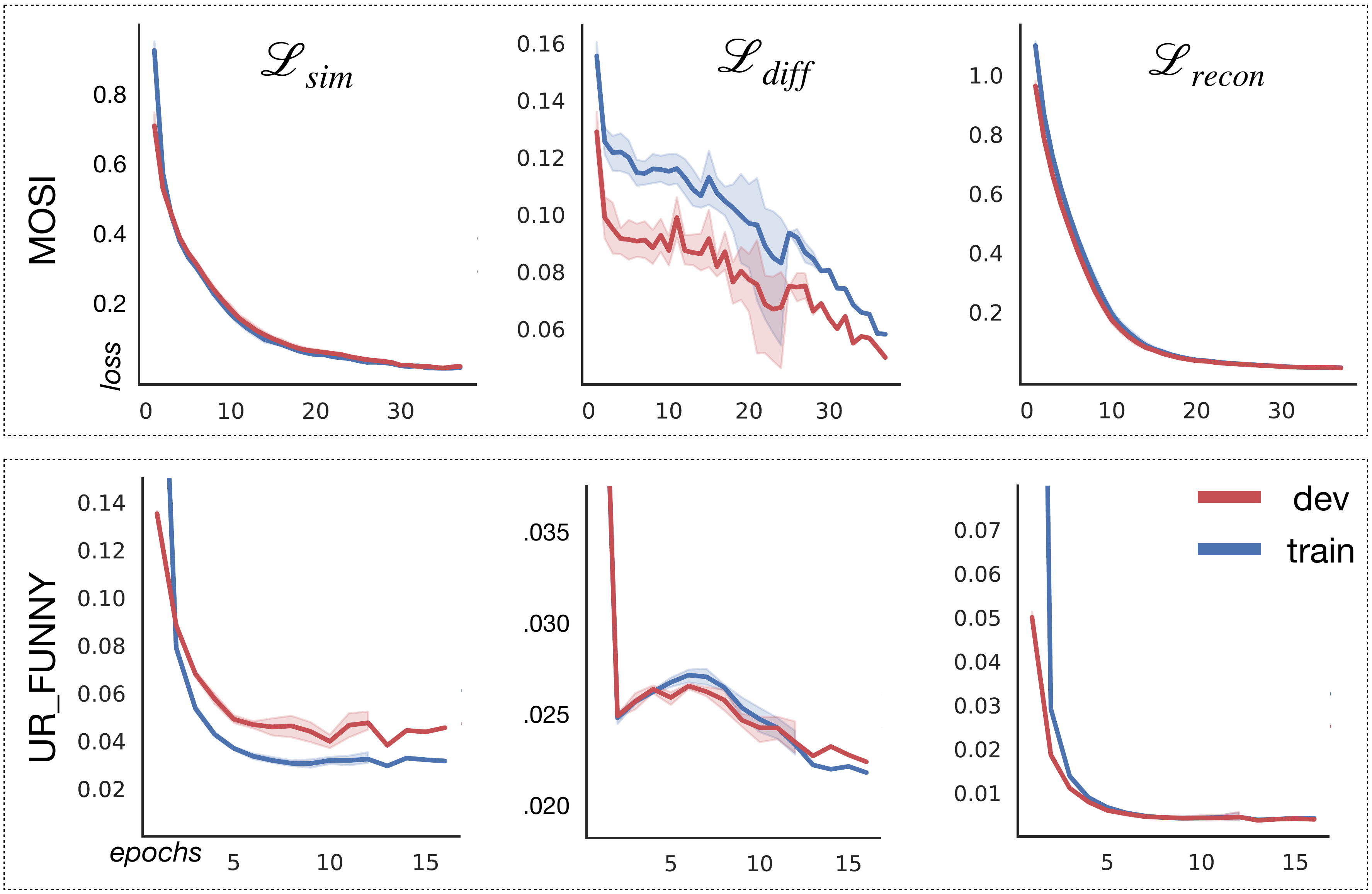}
    \caption{Trends in the regularization losses as training proceeds (values are for five runs across random seeds). Graphs depict losses in both training and validation sets for MOSI and UR$\_$FUNNY. Similar trends are also observed in MOSEI.}
    \label{fig:loss_trends}
\end{figure}

\paragraph{\textbf{Regularization Trends.}}
The losses $\{\mathcal{L}_{\text{sim}}, \mathcal{L}_{\text{diff}}, \mathcal{L}_{\text{recon}}\}$ act as measures to quantify how well the model has learnt modality-invariant and -specific representations. We thus trace the losses as training proceeds both in the training and validation sets. As seen in~\cref{fig:loss_trends}, all three losses demonstrate a decreasing trend with the number of epochs. This shows that the model is indeed learning the representations as per design. Like the training sets, the validation sets also demonstrate similar behavior.

\paragraph{\textbf{Visualizing Representations.}}

\begin{figure}[t]
    \centering
    \includegraphics[width=\linewidth]{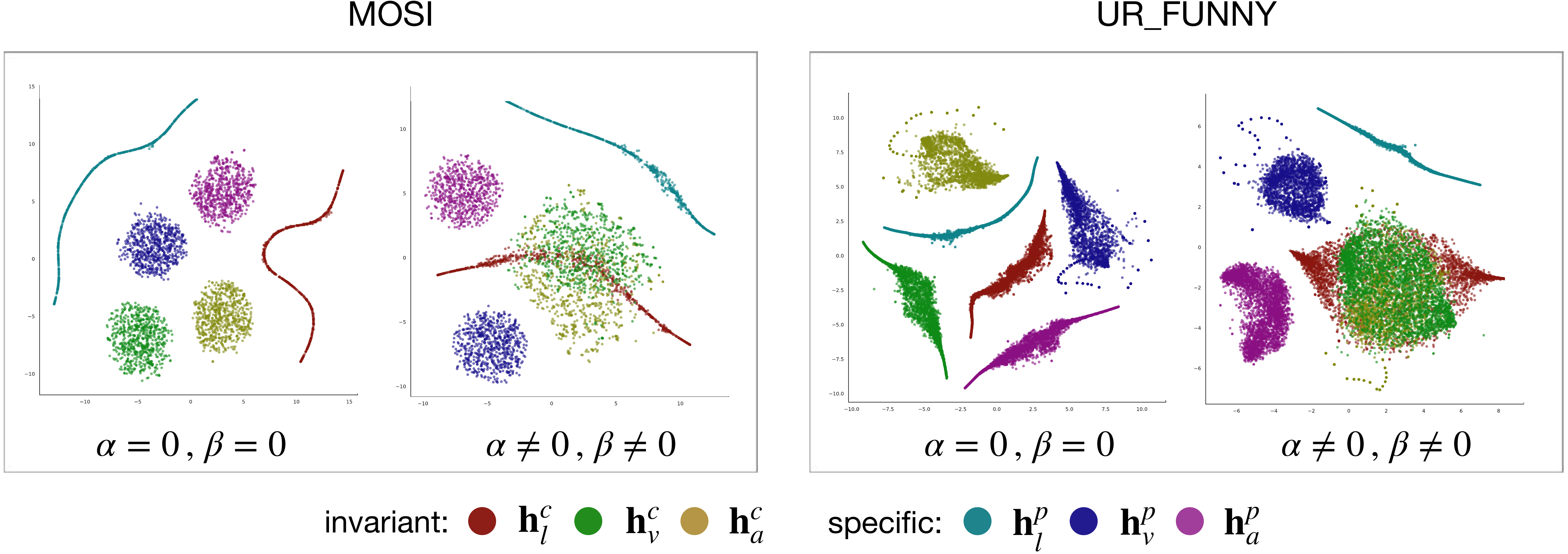}
    \caption{Visualization of the modality-invariant and -specific subspaces in the testing set of MOSI and UR$\_$FUNNY datasets using t-SNE projections~\cite{maaten2008visualizing}. Observations on MOSEI are also similar.}
    \label{fig:activations}
\end{figure}

While~\cref{fig:loss_trends} shows how regularization losses behave during training, it is also vital to investigate how well these characteristics are generalized. We thus visualize the hidden representations for the samples in the \textit{testing sets}. ~\cref{fig:activations} presents the illustrations, where it is clearly seen that in the case of no regularization ($\alpha = 0, \beta = 0$), modality-invariance is not learnt. Whereas, when losses are introduced, overlaps amongst the modality-invariant representations are observed. This indicates that MISA is able to perform desired subspace learning, even in the generalized scenario, i.e., in the testing set. We delve further into the utility of these subspaces in~\cref{sec:role_of_subspaces}.

\paragraph{\textbf{Importance of Regularization.}}

To quantitatively verify the importance of these losses, we take the best models in each dataset and re-train them by ablating one loss at a time. To nullify each loss, we set either $\{\alpha, \beta, \gamma\}$ to $0$. Results are observed in~\cref{tab:ablation} (model 5,6,7). As seen, the best performance is achieved when all the losses are involved. In a closer look, we can see that the models are particularly sensitive to the \textit{similarity} and \textit{difference} losses that ensures both the modality invariance and specificity. This dependence indicates that having separate subspaces is indeed helpful. For the \textit{reconstruction} loss, we see a lesser dependence on the model. One possibility is that, despite the absence of reconstruction loss,  the modality-specific encoders are not resorting to trivial solutions and rather learning informative representations using the task loss. This would not be the case if only the modality-invariant features were used for prediction.

\subsubsection{\textbf{Role of subspaces.}} \label{sec:role_of_subspaces}

In this section, we look at several variants to our proposed model to investigate alternative hypotheses:
\begin{enumerate}[1),leftmargin=*]
    \item \textit{MISA-base} is a baseline version where we do not learn disjoint subspaces. Rather, we utilize three separate encoders for each modality --- similar to previous works --- and employ fusion on them.
    \item \textit{MISA-inv} is a variant where there is no modality-specific representation. In this case, only modality-invariant representations are learnt and subsequently utilized for fusion.
    \item The next two variants, \textit{MISA-sFusion} and \textit{MISA-iFusion} are identical to MISA in the representation learning phase. In \textit{MISA-sFusion}, we only use the modality-specific features ($\mathbf{h}_{\{l/v/a\}}^{p}$) for fusion and prediction. Similarly, \textit{MISA-iFusion} uses only modality-invariant features ($\mathbf{h}_{\{l/v/a\}}^{c}$) for fusion.
\end{enumerate}

We summarize the results in~\cref{tab:ablation} (model 8-11). Overall, we find our final design to be better than the variants. Amongst the variants, we observe that learning only an invariant space might be too restrictive as not all modalities in an utterance share the same polarity stimulus. This is reflected in the results where \textit{MISA-inv} does not fare better than the general \textit{MISA-base} model. Both \textit{MISA-sFusion} and \textit{-iFusion} improve the performances but the best combination is when both representation learning and fusion utilize both the modality subspaces, i.e., the proposed model MISA.  

\subsubsection{\textbf{Visualizing Attention}}

\begin{figure}[t]
    \centering
    \includegraphics[width=\linewidth]{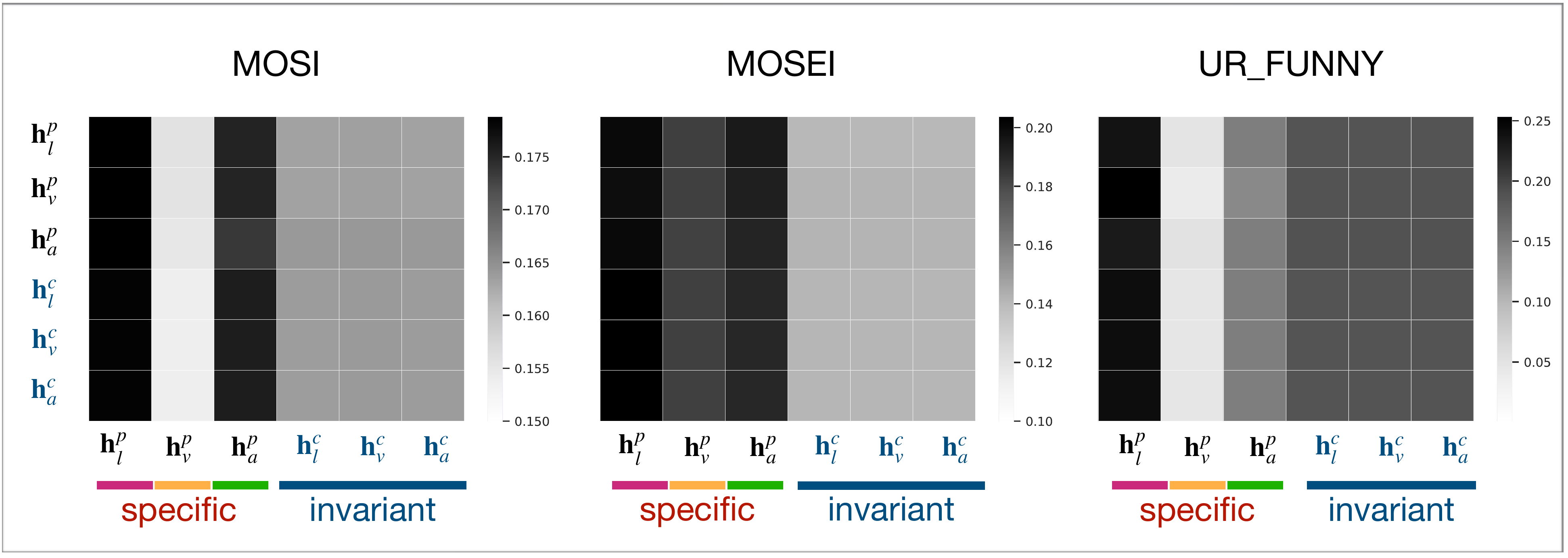}
    \caption{Average self-attention scores from the Transformer-based fusion module. The rows depict the \textit{queries}, columns depict the \textit{keys} (see~\cref{sec:fusion}). Essentially, each column represents the contribution of an input feature vector $\in \{\mathbf{h}^c_l, \mathbf{h}^c_v, \mathbf{h}^c_a, \mathbf{h}^p_l, \mathbf{h}^p_v, \mathbf{h}^p_a\}$ to generate the output feature vectors $[\mathbf{\bar{h}}^c_l, \mathbf{\bar{h}}^c_v, \mathbf{\bar{h}}^c_a, \mathbf{\bar{h}}^p_l, \mathbf{\bar{h}}^p_v, \mathbf{\bar{h}}^p_a]$.}
    \label{fig:attention}
\end{figure}

To analyze the utility of the learned representations, we look at their role in the fusion step. As discussed in~\cref{sec:fusion}, fusion includes a self-attention procedure on the modality representations that enhances each representation $\mathbf{h}^{c/p}_{l/v/a} $ to $\mathbf{\bar{h}}^{c/p}_{l/v/a}$, using a soft-attention combination of all its fellow representations (including itself). \cref{fig:attention} illustrates the average attention distribution of the testing sets. Each row in the figure is a probability distribution for the respective representation (averaged over all the testing samples). Looking at the columns, each column can be seen as the contribution that any vector $\mathbf{h} \in \{\mathbf{h}^{c/p}_{l/v/a}\}$ has to the enhanced representations of all the resulting vectors $\mathbf{\bar{h}}^{c/p}_{l/v/a}$. We observe two important patterns in the figures. First, we notice that the invariant representations influence equally amongst all three modalities. This is true for all the datasets and expected as they are aligned in the shared space. It also establishes that modality gap is reduced amongst the invariant features. Second, we notice a significant contribution from modality-specific representations. Although the average importance of a modality depends on the dataset, language, as also seen in quantitative results, contributes the most while acoustic and visual modalities provide varied levels of influences. Nevertheless, both invariant and specific representations provide information in the fusion, as observed in these influence maps.

\section{Conclusion} \label{sec:conclusion}

In this paper, we presented MISA -- a multimodal affective framework that factorizes modalities into modality-invariant and modality-specific features and then fuses them to predict affective states. Despite comprising simple feed-forward layers, we find MISA to be highly effective and observe significant gains over state-of-the-art approaches in multimodal sentiment analysis and humor detection tasks. Explorative analysis reveals desirable traits, such as reductions in modality gap, being learned by the representation learning functions, which obviates the need for complex fusion mechanism. Overall, we stress the importance of representation learning as a pre-cursory step of fusion and demonstrate its efficacy through rigorous experimentation. The codes for our experiments are available at: \url{https://github.com/declare-lab/MISA}.

In the future, we plan to analyze MISA in other dimensions of affect, such as emotions. Additionally, we also aim to combine the MISA framework with other fusion schemes to try for further improvements. Finally, the similarity and difference loss modeling allow various metrics and regularization choices. We thus intend to analyze other options in this regard.

\begin{acks}
This research is supported by ($i$) Singapore Ministry of Education Academic Research Fund Tier 2 under MOE's official grant number MOE2018-T2-1-103, ($ii$) A*STAR under its RIE 2020 Advanced Manufacturing and Engineering (AME) programmatic grant, Award No. -  A19E2b0098, and ($iii$) DSO, National Laboratories, Singapore under grant number RTDST190702 (Complex Question Answering).
\end{acks}

\newpage

\bibliographystyle{ACM-Reference-Format}
\bibliography{references}


\appendix

\section{Baseline Models} \label{sec:appendix_baseline_models}
This section provides the details of the baseline models mentioned in the paper.

\begin{itemize}[leftmargin=*]
    \item \textbf{MFN}~\cite{DBLP:conf/aaai/ZadehLMPCM18}~\footnote{\url{https://github.com/pliang279/MFN}} is a multi-view gated memory network that stores intra- and cross-view interactions in its memories.
    \item \textbf{RAVEN}~\cite{DBLP:conf/aaai/WangSLLZM19} utilizes an attention-based model on non-verbal signals to re-adjst word embeddings based on the multimodal context.
    \item \textbf{MARN}~\cite{DBLP:conf/aaai/ZadehLPVCM18} learns intra-modal and cross-modal interactions by designing a hybrid LSTM memory component.
    \item \textbf{MV-LSTM}~\cite{DBLP:conf/eccv/RajagopalanMBG16} proposes a multi-view LSTM variant with designated representations for each view inside the LSTM function.
    \item \textbf{RMFN}~\cite{DBLP:conf/emnlp/LiangLZM18} decomposes the fusion process into multi-stage computations, where each stage focuses on a subset of multimodal signals.
    \item \textbf{Graph-MFN}~\cite{DBLP:conf/acl/MorencyCPLZ18} is a development over the MFN fusion model which adds a dynamic graph module on top of its recurrent structure. The nodes of the graph are the various interactions (bi-modal, tri-modal) with a hierarchical topology.
    \item \textbf{TFN}~\cite{DBLP:conf/emnlp/ZadehCPCM17}~\footnote{\url{https://github.com/Justin1904/TensorFusionNetworks}} calculates a multi-dimensional tensor (based on outer-product) to capture uni-, bi-, and tri-modal interactions.
    \item \textbf{LMF}~\cite{DBLP:conf/acl/MorencyLZLSL18}~\footnote{\url{https://github.com/Justin1904/Low-rank-Multimodal-Fusion}} is an improvement over TFN, where low-rank modeling of the TFN tensor is proposed.
    \item \textbf{MFM}~\cite{DBLP:conf/iclr/TsaiLZMS19}~\footnote{\url{https://github.com/pliang279/factorized/}} learns discriminative and generative representations for each modality where the former is used for classification while the latter is used to learn the modality-specific generative features.
    \item \textbf{LMFN}~\cite{DBLP:journals/tmm/MaiXH20} segments the utterance vectors from each modality into smaller sections and performs fusion in these local regions, followed by global fusion across these fused segments.
    \item \textbf{HFFN}~\cite{DBLP:conf/acl/MaiHX19} follows a similar strategy where the local fusion is termed as \textit{divide}, and \textit{combine}, while the global fusion is termed as \textit{conquer} phase.
    \item \textbf{MulT}~\cite{DBLP:conf/acl/TsaiBLKMS19} proposes a multimodal transformer architecture which translates one modality to another using directional pairwise cross-attention.
    \item \textbf{MCTN}~\cite{DBLP:conf/aaai/PhamLMMP19} implements a translation-based model with an encoder-decoder setup to convert one modality to another. Coupled with cyclic consistency losses, the encoding representation learns informative common representations from all modalities.
    \item \textbf{ARGF}~\cite{DBLP:journals/corr/abs-1911-07848} is a recent model that learns a common embedding space by translating a source modality to a target modality through adversarial training, and maked predictions using graph-based fusion mechanism.
\end{itemize}

From the contextual-utterance family of works, we consider the following baseline models:

\begin{itemize}[leftmargin=*]
    \item \textbf{BC-LSTM}~\cite{DBLP:conf/acl/PoriaCHMZM17} is a contextual-utterance model which learns an bi-directional LSTM model overall the whole video, thus framing the problem as a structured prediction task.
    \item \textbf{CH-FUSION}~\cite{DBLP:journals/kbs/MajumderHGCP18} is a strong baseline which performs hierarchical fusion by composing bi-modal interactions followed by tr-modal fusion.
    \item  \textbf{CIM-MTL}~\cite{DBLP:conf/naacl/AkhtarCGPEB19} is a multi-task version of MMMU-BA, analyzing the sentiment as well as emotion to exploit finds the inter-dependence between them.
    \item \textbf{CIA}~\cite{DBLP:conf/emnlp/ChauhanAEB19} is another contextual model whose core functionality is to learn cross-modal auto-encoding to learn modality translations and utilize this feature in a contexual attention framework.
    \item  \textbf{DFF-ATMF}~\cite{chen2019complementary} is a bi-modal model which first learns individual modality features followed by attention-based modality fusion.
    \item \textbf{C-MFN}~\cite{DBLP:conf/emnlp/HasanRZZTMH19} is the present state-of-the-art for UR$\_$FUNNY. Essentially it is based on the MFN fusion mechanism and extends it to the contextual regime which takes the previous sequence of utterances into account along with MFN-style multimodal fusion.
\end{itemize}

\section{Dataset Sizes} \label{sec:appendix_sizes}

\cref{tab:dataset_stats} provides the sizes (number of utterances) in each dataset. 

\begin{table}[h]
  \centering
  \resizebox{0.8\linewidth}{!}{
  \begin{tabular}{|l|c:c:c|}
    \hline
    Datasets & MOSI & MOSEI & UR$\_$FUNNY\\
    mode & $\#$utterances & $\#$utterances & $\#$utterances \\
    \hline
    \hline
    train & 1283 & 16315 & 10598 \\
    dev & 229 & 1871 & 2626 \\
    test & 686 & 4654 & 3290 \\
    \hline
  \end{tabular}
  }
  \caption{Sizes of the datasets.}
  \vspace{-0.5cm}
  \label{tab:dataset_stats}
\end{table}

\section{Hyper-parameter Selection} \label{sec:hyperparams}

To select appropriate hyper-parameters, we utilize the validation sets provided in the datasets. We perform grid-search over the hyper-parameters to select the model with best validation classification/regression loss. We look over finite sets of options for hyper-parameters. These include non-linear activations: \textit{leakyrelu}\footnote{\protect\url{https://pytorch.org/docs/stable/nn.html\#leakyrelu}}, \textit{prelu}\footnote{\url{https://pytorch.org/docs/stable/nn.html\#prelu}}, \textit{elu}\footnote{\url{https://pytorch.org/docs/stable/nn.html\#elu}}, \textit{relu}\footnote{\url{https://pytorch.org/docs/stable/nn.html\#relu}}, and \textit{tanh}\footnote{\url{https://pytorch.org/docs/stable/nn.html\#tanh}}, $\alpha \in \{0.7, 1.0\}$, $\beta \in \{0.3, 0.7, 1.0\}$, and $\gamma \in \{0.7, 1.0\}$. Finally, we look at dropout values from $\{0.1, 0.5, 0.7\}$. For optimization, we utilize the Adam optimizer with an exponential decay learning rate scheduler. The training duration of each model is governed by early-stopping strategy with a patience of $6$ epochs. The final hyper-parameters for each model per dataset is summarized in~\cref{tab:hyper-parameters}.

\begin{table}[h]
  \centering
  \resizebox{0.8\linewidth}{!}{
  \begin{tabular}{|l|c|c|c|}
    \hline
    Hyper-param & \multicolumn{1}{c}{MOSI}&\multicolumn{1}{c}{MOSEI}&\multicolumn{1}{c|}{UR$\_$FUNNY}\\
    \hline
    \hline
    cmd K & 5 & 5 & 5 \\
    activation &  ReLU & LeakyReLU & Tanh \\
    batch size &  64 & 16 & 32 \\
    gradient clip &  1.0 & 1.0 & 1.0 \\
    $\alpha$ & 1.0 & 0.7 & 0.7 \\
    $\beta$ & 0.3 & 0.3 & 1.0 \\
    $\gamma$ & 1.0 & 0.7 & 1.0 \\
    dropout & 0.5 & 0.1 & 0.1 \\
    $d_h$ & 128 & 128 & 128 \\
    learning rate & 1e-4 & 1e-4 & 1e-4 \\
    \hline
  \end{tabular}
  }
  \caption{Final hyper-parameter values in each dataset.}
  \vspace{-0.5cm}
  \label{tab:hyper-parameters}
\end{table}

\section{Network Topology} \label{sec:topology}

\begin{figure*}[ht]
\begin{subfigure}{.8\textwidth}
  \centering
  \includegraphics[width=.8\linewidth]{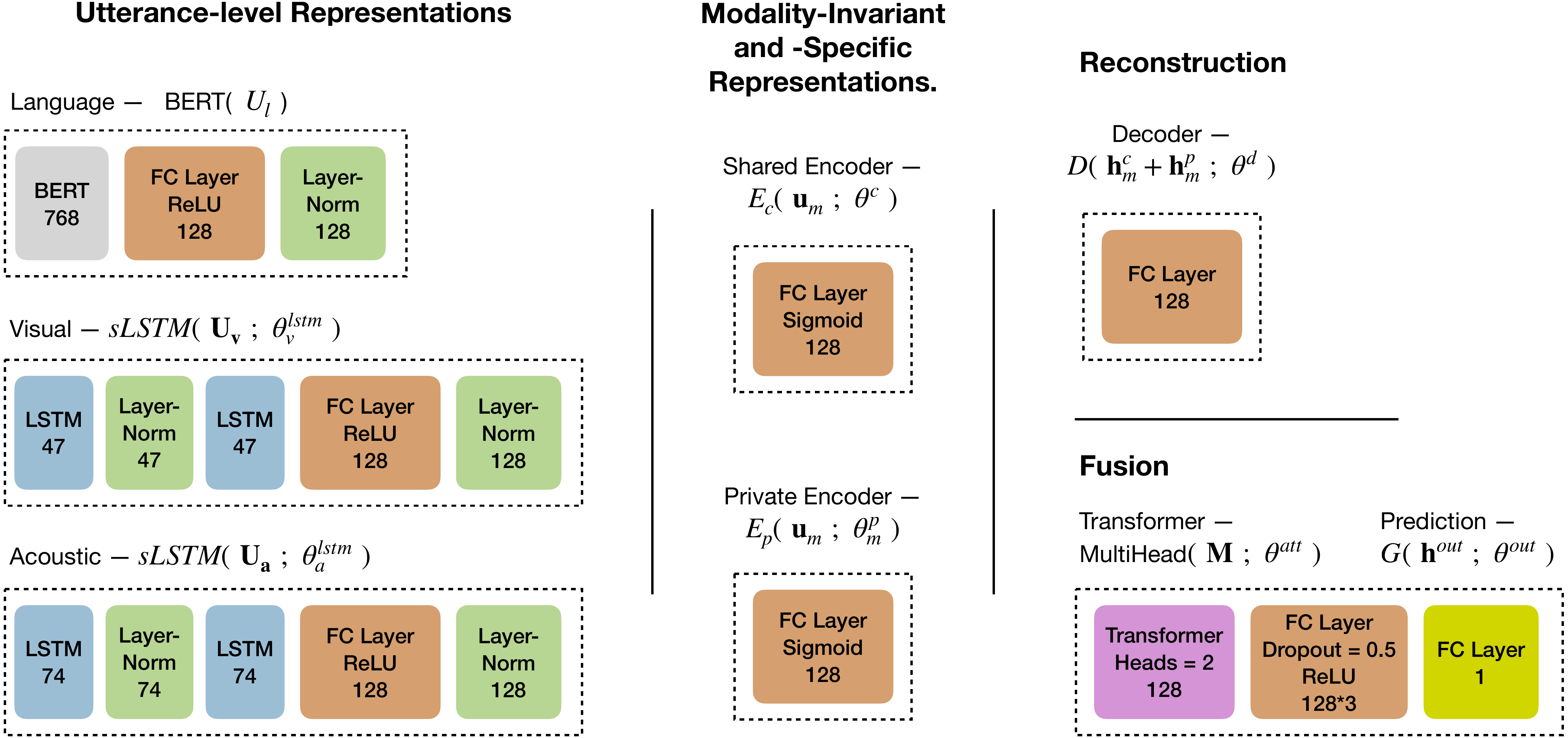}  
  \caption{MOSI}
\end{subfigure}

\vspace{0.5cm}

\begin{subfigure}{.8\textwidth}
  \centering
  \includegraphics[width=.8\linewidth]{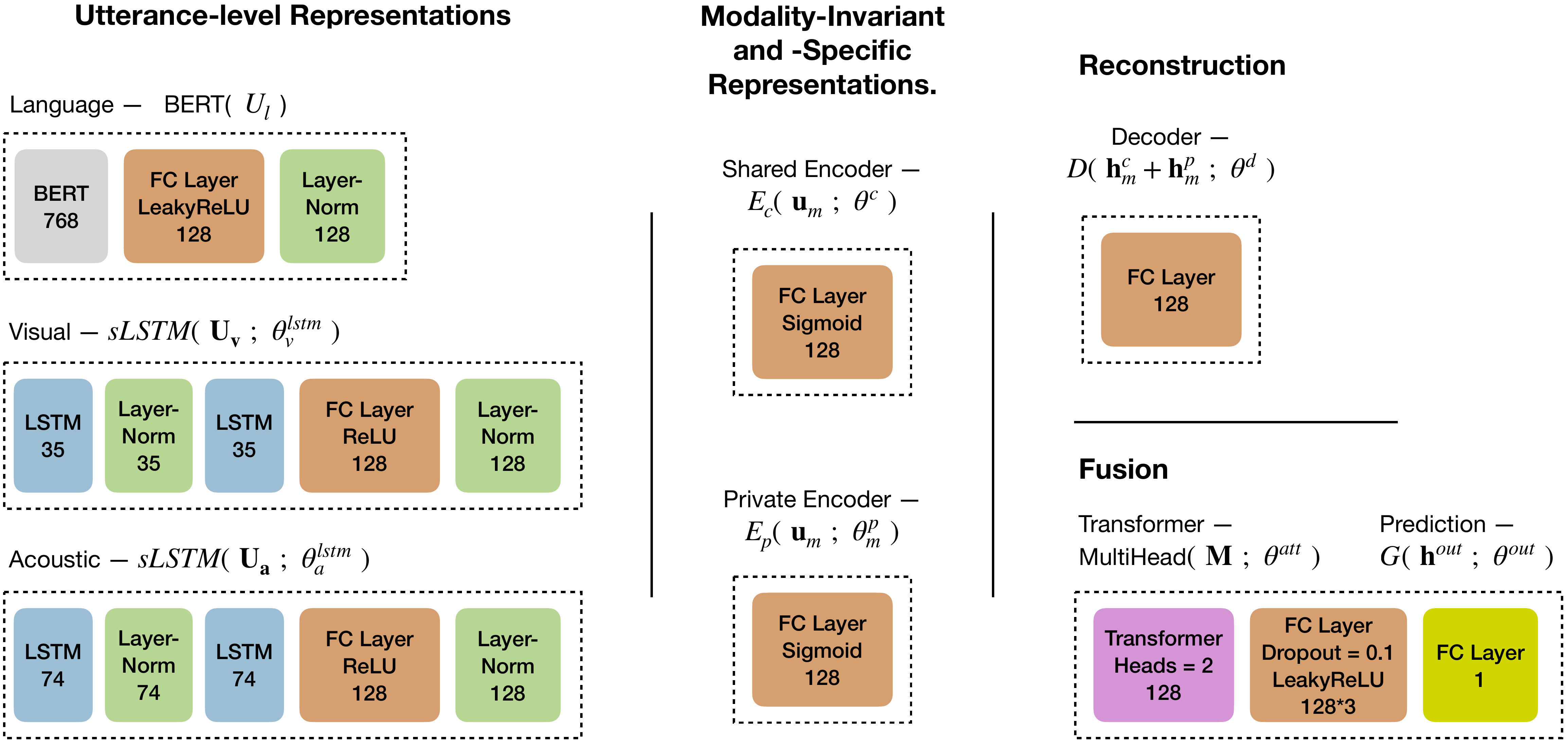}  
  \caption{MOSEI}
\end{subfigure}

\vspace{0.5cm}

\begin{subfigure}{.8\textwidth}
  \centering
  \includegraphics[width=.8\linewidth]{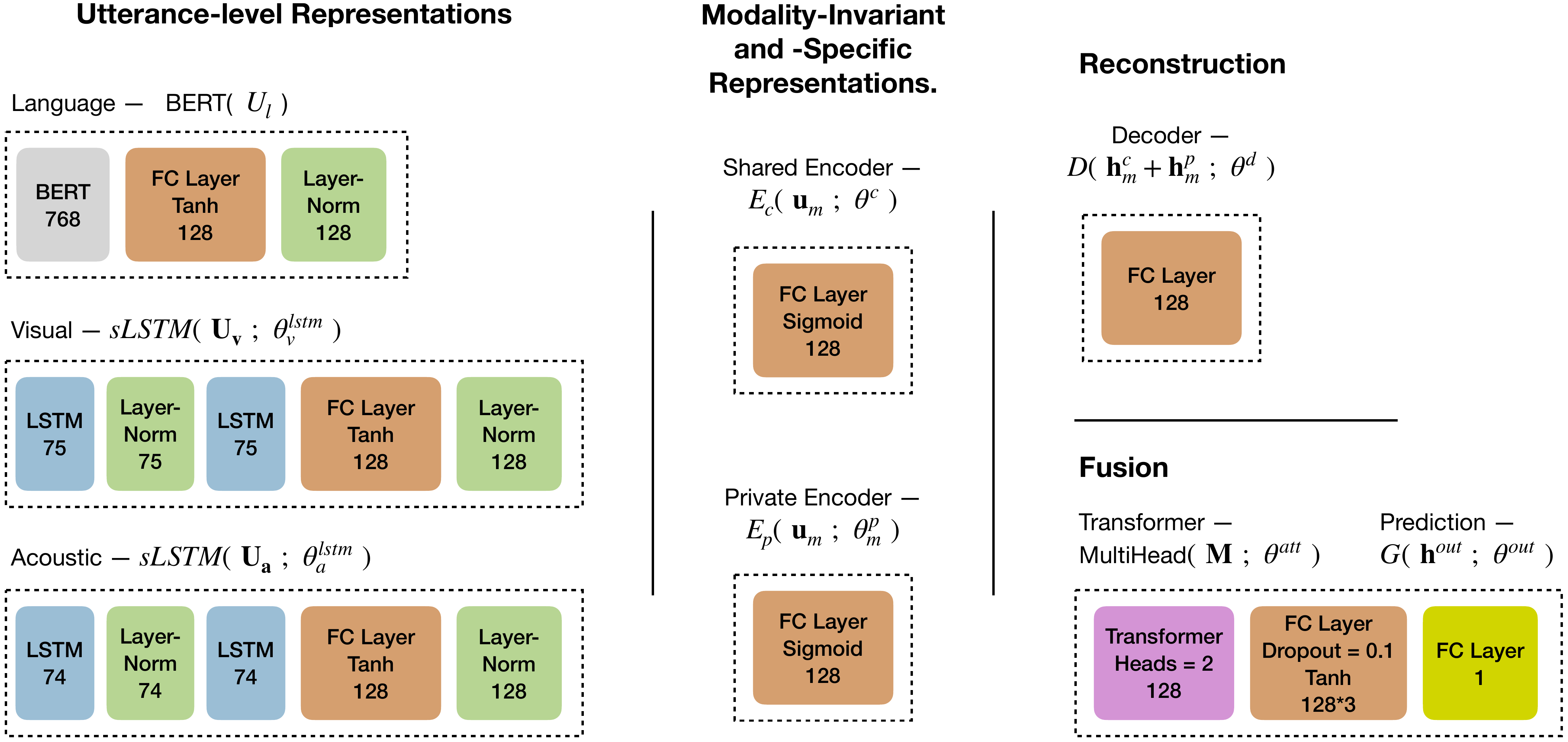}  
  \caption{UR$\_$FUNNY}
\end{subfigure}

\caption{Description of the topologies used for the different datasets.}
\label{fig:topology}
\end{figure*}

\cref{fig:topology} describes the network topologies of the final models used in each dataset.

\end{document}